\pdfoutput=1

\documentclass[11pt]{article}

\usepackage[]{emnlp2021}

\usepackage{times}
\usepackage{latexsym}

\usepackage[T1]{fontenc}

\usepackage[utf8]{inputenc}

\usepackage{microtype}

\usepackage{amsmath}
\usepackage{booktabs}
\usepackage{bm}
\usepackage{enumitem, amsmath, amsthm, epsfig, graphicx, balance, subfig, multirow, amssymb, mathrsfs}
\usepackage{xcolor, color}
\usepackage{diagbox}
\usepackage{multirow}
\usepackage{booktabs}

\newtheorem{definition}{\noindent Definition}
\newcounter{example}[section]
\newenvironment{example}[1][]{\refstepcounter{example}\par\medskip
   \noindent \textbf{Example~\theexample. #1} \rmfamily}{\medskip}
\usepackage{xspace}
\newcommand{\our}{\mbox{HashNews}\xspace}
\newcommand\blfootnote[1]{%
  \begingroup
  \renewcommand\thefootnote{}\footnote{#1}%
  \addtocounter{footnote}{-1}%
  \endgroup
}

\title{News Meets Microblog: Hashtag Annotation via Retriever-Generator}

\author{
Xiuwen Zheng$^{1,2\dagger}\;\;\;\;\;$ Dheeraj Mekala$^{1\dagger}\;\;\;\;\;$ Amarnath Gupta$^{2}\;\;\;\;\;$ Jingbo Shang$^{1,3}$ \\
\small $^1$ Department of Computer Science and Engineering, University of California San Diego, CA, USA \\
\small $^2$ San Diego Supercomputer Center, University of California San Diego, CA, USA \\
\small $^3$ Hal\i c\i o\u glu Data Science Institute, University of California San Diego, CA, USA \\
\small \texttt{\{xiz675, dmekala, a1gupta, jshang\}@ucsd.edu}
}

\begin{document}
\maketitle
\begin{abstract}
 \blfootnote{$\dagger$ Represents equal contribution}
Hashtag annotation  for microblog posts has been recently formulated as a sequence generation problem to handle emerging hashtags that are unseen in the training set. The state-of-the-art method leverages conversations initiated by posts to enrich contextual information for the short posts. 
However, it is unrealistic to assume the existence of conversations before the hashtag annotation itself. 
Therefore, we propose to leverage news articles published before the microblog post to generate hashtags following a Retriever-Generator framework.
Extensive experiments on English Twitter datasets demonstrate superior performance and significant advantages of leveraging news articles to generate hashtags. 
\end{abstract}
\section{Introduction}
Hashtag annotation, which aims to generate hashtags for microblog posts, is a fundamental task in microblog platforms (e.g., Twitter) because hashtags play a crucial role in user engagement and a wide range of downstream tasks~\cite{zhang2019hashtag}.
This problem was recently formulated as sequence generation, so it can handle emerging hashtags that never or rarely appeared in training. 
Our analysis on 36K tweets shows that more than $80\%$ hashtags appear at most 5 times (Appendix A), verifying the necessity of this formulation.

Microblog posts are usually short, which makes it hard to generate hashtags merely from the posts.
State-of-the-art method~\cite{wang2019microblog} introduces conversations initiated by the post for more context.
However,
such conversations are not available when composing the post and hashtags.

\begin{table}[ht]
    \centering
    \includegraphics[width=0.85\linewidth]{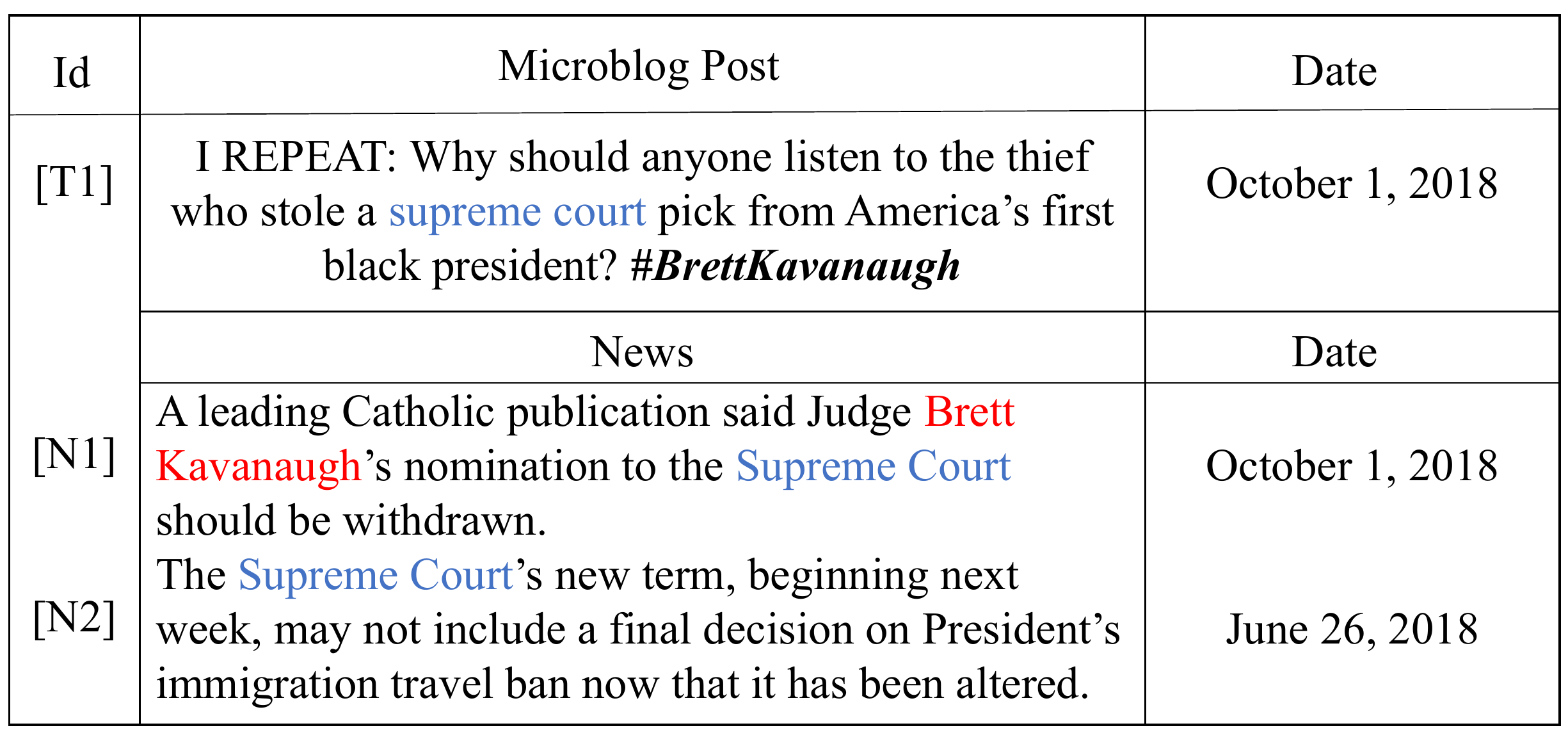}
    \vspace{-3mm}
    \caption{A microblog post and its related news articles. Entities based on which news are retrieved are in \textcolor{blue}{blue.} Words indicative for the hashtag are in \textcolor{red}{red.}} \label{figure:intro-ex}
    \vspace{-5mm}
\end{table}

\begin{figure*}
    \centering
    \includegraphics[width=\textwidth]{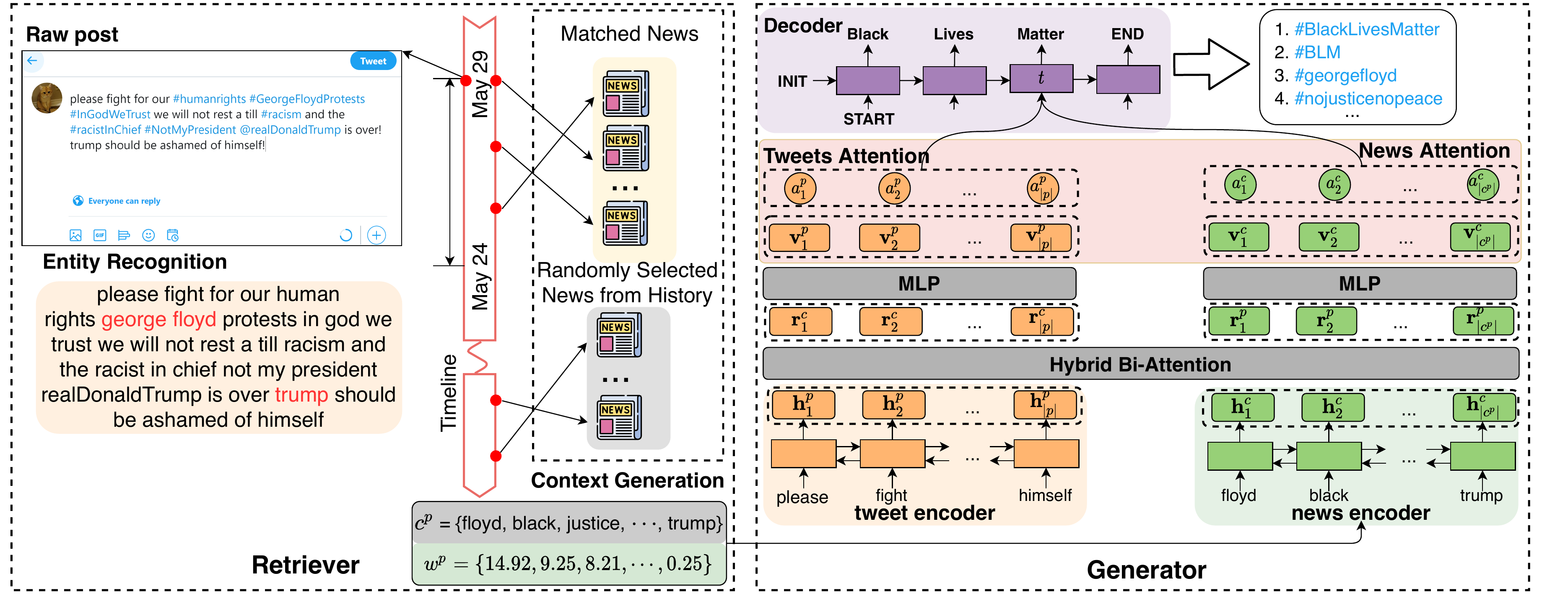}
    \vspace{-0.7cm}
    \caption{Overview of our Retriever-Generator framework.}
    \vspace{-0.3cm}
    \label{figure:model}
\end{figure*}

In fact, microblog posts are typically reflections of  recent events happening in the real world and the information of such events would be widely available in news articles.
Therefore, news articles could serve as a powerful source for hashtag annotation in addition to the post itself.
In our experiments, we observe that around 28\% of named entities in hashtags are present in retrieved news articles.
Table~\ref{figure:intro-ex} presents an example where a microblog post $T1$ and news $N1$ talk about the same event (i.e., choice of supreme court judge).
The hashtag \textit{\#BrettKavanaugh} is not present in the post, however, it appears in $N1$.
We also observe that recent news articles are more relevant to the post content than older news articles.
For example, the news $N2$ talks about the supreme court, however, it is outdated to predict the hashtags. 

Inspired by these observations,
we propose \our following a Retriever-Generator framework as illustrated in Figure~\ref{figure:model}.
Our retriever adapts the traditional information retrieval model to emphasize emerging entities while retrieving relevant news articles.
Our generator jointly models posts and relevant news with a novel hybrid bi-attention. 

Our contributions are summarized as follows:
\begin{itemize}[nosep, leftmargin=*]
    \item We develop a time-aware entity-focused ranking function that ranks news articles by taking temporal popularity of entities into consideration.
    \item We design a novel hybrid bi-attention that captures effects of news and posts on each other with emphasis on important entities from news.
    \item Experimental results on two real-world datasets demonstrate superiority of leveraging news articles to generate hashtags.
\end{itemize}
To the best of our knowledge, this is the first work on leveraging news articles for hashtag annotation with a sequence generation architecture.

\noindent\textbf{Reproducibility.} Our code is made public here\footnote{\url{https://github.com/dheeraj7596/HashNews}} and datasets will be provided upon request. 
\vspace{-0.3cm}

\section{Our Retriever-Generator Framework}
\vspace{-0.3cm}
Given a microblog post $p$ and a set of news articles  published 
before $p$, our goal is to generate hashtag $y$ represented by a word sequence $\langle y_1, y_2, \cdots y_{|y|} \rangle$.
Here, $|y|$ is the number of the words.

We propose a Retriever-Generator framework shown in  
Figure~\ref{figure:model}. 
It has two modules: (1) a time-aware entity-focused news retriever and (2) a hybrid bi-attention hashtag generator.




\subsection{Time-Aware Entity-Focused Retriever}
\begin{table*}[t]
\centering 
\caption{Statistics of our datasets. \textbf{\% of entity hashtags}: the number of hashtags with at least one entity, \textbf{\% of all-entity hashtags}: the number of hashtags consisting of only entities, $\bm{P^p}$: percentage of words in hashtags that are also present in their corresponding posts.}
\vspace{-3mm}
\scalebox{0.9}{
\small
\vspace{-3mm}
\label{tab:statistics}
\begin{tabular}{cccccccc} 
    \toprule
    \textbf{Datasets}   & 
    \begin{tabular}[c]{@{}c@{}}\textbf{\# of}\\\textbf{posts}\end{tabular} &
    \begin{tabular}[c]{@{}c@{}}\textbf{\# of distinct}\\\textbf{hashtags}\end{tabular} &
    \begin{tabular}[c]{@{}c@{}}\textbf{Avg \# of hashtags}\\\textbf{per post}\end{tabular} &
    \begin{tabular}[c]{@{}c@{}}\textbf{Avg \# of entities}\\\textbf{per post}\end{tabular} &
    \begin{tabular}[c]{@{}c@{}}\textbf{\% of entity}\\\textbf{hashtags}\end{tabular} & \begin{tabular}[c]{@{}c@{}}\textbf{\% of all-entity}\\\textbf{hashtags}\end{tabular} &
    \begin{tabular}[c]{@{}c@{}}$\bm{P^p}$\end{tabular}\\ 
    \midrule
    Tweets2018 & 35966 & 19635 & 2.70 & 1.78 & 40.48 & 21.43 & 9.49\\ 
    Tweets2020 & 27418 & 14687 & 2.24 & 1.02 & 47.47 & 18.81 & 11.79\\ \bottomrule
\end{tabular}
}
\vspace{-5mm}
\end{table*}


Entities, especially emerging ones, shall be more important than other words to analyze the focus of a post. 
As shown in Table~\ref{tab:statistics}, we observe that more than $40\%$ hashtags in our datasets contain at least one entity, and about $20\%$ hashtags consist of only entities.
Thus, we extract entities in a post as query keywords to retrieve relevant news. 
During retrieval, we emphasize important entities by exploiting their temporal popularities.  

\noindent\textbf{Temporal Popularity of Entities.}
We propose an adaptive accumulation method to collect candidate news articles based on timestamps. Specifically, for a post $p$ created at day $t$, we construct a series of $k$ candidate corpora ($k=5$ in experiments) $\mathcal{D}$ = $\{D_1, D_2, \ldots, D_{k-1}, D_k\}$.
$D_i$ contains all the news articles published in the time window from day $t-i$ to day $t$. 
From each corpus $D_i \in \mathcal{D}$, we retrieve the most relevant article (excluding the ones have been retrieved from $D_1, \ldots, D_{i-1}$), leading to a total of $k$ retrieved news articles. 
$D_i$ is a subset of $D_{i+1}$, so, all $k$ news articles could be from $D_1$.

Given an entity $e$ and a candidate corpus $D_i$, we compute the temporal popularity of $e$ through a comparative analysis.
Intuitively, if an entity becomes more popular in the recent news than usual, it can be referred as an emerging entity, which should be emphasised more when ranking news articles.
Therefore, we randomly sample a time-independent news corpus $R$ and compare the inverse document frequency (IDF)~\cite{han2011data} values of entity $e$ in the two corpora, $D_i$ and $R$.
Formally, we define the temporal popularity (TP) of $e$ in corpus $D_i$ as 

\vspace{-0.3cm}
$$
\footnotesize
\begin{aligned}
    \text{TP}(e, D_i) = \frac{IDF(e, R)}{IDF(e, D_i)}.
\end{aligned}
$$

\noindent\textbf{Ranking Function.}
Our ranking function is inspired by the Okapi BM25 function \cite{robertson1995okapi} and further incorporates temporal popularity to emphasize emerging entities.
Given a microblog post $p$ and a news  $d$ from the corpus $D_i$, the score is calculated as follows:
$$
\footnotesize
  \begin{aligned}
    s(p, d | D_i)=\sum_{e \in p} \text{TP}(e, D_i) {\frac{f_{e}^{d} \cdot (a+1)}{f_{e}^{d} + a \left(1 - b + b \cdot {\frac {|d|}{\mathbb{E}_{D_i}[|d|]}}\right)}}
\end{aligned}  
$$
where $f_{e}^{d}$ is the frequency of entity $e$ in the document $d$. $\mathbb{E}_{D_i}[|d|]$ is the average document length of corpus $D_i$. $a$ and $b$ are parameters inherent from Okapi BM25 typically set to 1.2 and 0.75 \cite{jones2000probabilistic}.

Microblogs are usually written in an informal style, so two different phrases in a post and  a news article may refer to the same entity in the real world.
Therefore, the entities in posts and news have to be carefully matched to better estimate $f_{e}^{d}$ and $IDF(e,\cdot)$. We develop a soft entity matching mechanism described in detail in Appendix B. 



For each microblog $p$, we rank news articles in $\mathcal{D}$ using the above ranking function and retrieve top-$k$ articles $H = \{h_1, h_2, \ldots, h_k\}$.
Since news articles are usually long, to reduce computational cost,
we construct context words $c^p$ and use them as input to generator. The weight of a word $t$ in selected news is computed by,

\vspace{-0.3cm}
\begin{equation}\label{eq:weight}
\footnotesize
\begin{aligned}
    w^p_t = \sum_{i=1}^{k} s(p, h_i | D_i) \cdot freq(t, h_i)
\end{aligned}
\end{equation}
where $freq(t, h_i)$ denotes the frequency of $t$ in news $h_i$. Words with the highest weights are selected as context words $c^p$ for post $p$.

\vspace{-0.2cm}
\subsection{Hybrid Bi-Attention Generator}
\vspace{-0.2cm}
Our generator extends the dual encoder network~\cite{wang2019microblog} which consists of two encoders, one for the microblog post and the other for the context constructed from news as illustrated in Figure~\ref{figure:model}.  
To effectively capture their joint effects on each other, we employ a novel hybrid bi-attention over the encoders’ outputs, incorporating local weights of context tokens from Eq.~\eqref{eq:weight}.

\noindent\textbf{Encoders.}
The inputs for the two encoders are post tokens  $\mathbf{x}^p_1,\cdots,\mathbf{x}^p_{|p|}$ and context tokens  $\mathbf{x}^c_1,\cdots,\mathbf{x}^c_{|c^p|}$.
We adopt the bidirectional gated recurrent unit (Bi-GRU) \cite{cho2014learning} for both encoders to encode a post token $\mathbf{x}^p_i$ into $\mathbf{h}_i^p$ and context token $\mathbf{x}^c_i$ into $\mathbf{h}_i^c$. 
Specifically, $\mathbf{h}_i^p=[\overrightarrow{\mathbf{h}_i^p}; \overleftarrow{\mathbf{h}_i^p}]$ is the concatenation of forward and backward hidden states for the $i$-th post token $\mathbf{x}^p_i$.
Likewise, $\mathbf{h}_j^c$ is the concatenation of forward and backward hidden states for the $j$-th context token $\mathbf{x}^c_j$ via another Bi-GRU.
$\mathbf{h}_i^p$ and $\mathbf{h}_j^c$ are $d$-dimensional vectors.
 




\noindent\textbf{Our Hybrid Bi-attention.}
Building upon traditional bi-attention~\cite{seo2016bidirectional}, we introduce local attentions to emphasize temporally popular  entities while obtaining news-aware post representation.
Our intuition is that a news article containing trending entities would get large ranking scores, and frequent entity tokens in these news will be selected out and assigned higher weights by the retriever (Eq.\eqref{eq:weight}).
Their weights should be taken into consideration while obtaining news-aware post representation.
Specifically, the  post-aware news representation $\mathbf{r}_j^p$  and news-aware post representation $\mathbf{r}_i^c$ are computed as follows,

\vspace{-0.5cm}
$$
\footnotesize
\begin{aligned}
    \mathbf{r}_j^p=\sum_{i=1}^{|p|}a_{ij}^p\cdot \mathbf{h}_i^p \quad \text{and}\quad\mathbf{r}_i^c=\sum_{j=1}^{|c^p|}a_{ij}^c\cdot \mathbf{h}_j^c
\end{aligned}
$$

\vspace{-0.3cm}
\noindent where

\vspace{-0.5cm}
$$
\footnotesize
    \begin{aligned}
        a_{ij}^p&=\frac{\exp(\mathbf{h}_i^p\cdot \mathbf{W}_b\cdot\mathbf{h}_j^c)}{\sum_{i'}\exp(\mathbf{h}_{i'}^p\cdot \mathbf{W}_b\cdot\mathbf{h}_{j}^c)},\\
        a_{ij}^c&=\frac{\exp(w^p_j\cdot \mathbf{h}_i^p\cdot \mathbf{W}_b\cdot\mathbf{h}_j^c)}{\sum_{j'}\exp(w^p_{j'}\cdot \mathbf{h}_{i}^p\cdot \mathbf{W}_b\cdot\mathbf{h}_{j'}^c)},\\
        w_j^p&=\text{weight of } j^{th} \text{ context token $\mathbf{x}^c_j$ by Eq.~\eqref{eq:weight}} .
    \end{aligned}
$$

\vspace{-0.3cm}
\noindent Note that, $\mathbf{W}_b\in\mathbb{R}^{d\times d}$ is a learnable weight matrix measuring the proximity between $\mathbf{h}_i^p$ and $\mathbf{h}_j^c$.  

\begin{table*}
\centering
\caption{Evaluation results of compared methods and our Retriever-Generator framework (\our).}\label{tab:main_result}
\vspace{-3mm}
\scalebox{0.85}{
\small
\begin{tabular}{c|cccccc|cccccc} 
\toprule
\multirow{2}{*}{\textbf{Models}} & \multicolumn{6}{c|}{\textbf{Tweets2018}} & \multicolumn{6}{c}{\textbf{Tweets2020}}                          \\ 
& $F_1@1$  & $F_1@5$  & $F_1@10$  & ACC    & MAP   & RG-1  & $F_1@1$  & $F_1@5$  & $F_1@10$  & ACC   & MAP  & RG-1   \\ 
\midrule
Seq2Seq                                           & 7.17     & 9.95     & 8.14      & 13.13 & 6.41  & 8.33  & 6.18     & 6.82     & 5.23      & 9.92  & 5.25 & 9.33   \\
Seq2Seq+Copy                                      & 7.97     & 11.13    & 8.92      & 14.51 & 7.21  & 9.18  & 6.81     & 7.45     & 5.69      & 10.73 & 5.64 & 9.67   \\ 
LSTM-TOP   &  10.18     &   14.03   &  11.36   &   18.65    & 8.36 &  8.92  & 9.65  &  \textbf{10.95}    &  \textbf{8.39}    &  15.51     &8.99  & 10.51  \\
\our                                         & \textbf{11.51}   & \textbf{15.16}   & \textbf{12.08}   & \textbf{21.11} & \textbf{10.96} & \textbf{13.36} & \textbf{10.59}    & 10.90 & 8.08  & \textbf{17.00} & \textbf{9.84} & \textbf{14.77}  \\
\midrule
\our-NoRank                                  &10.93& 14.50& 11.65 & 20.02 &10.20 &  12.69 & 9.26 &9.82 & 7.49& 14.92& 8.48&13.05\\
\our-NoRankNoLocal                           & 10.50    & 14.21    & 11.68     & 19.24 & 9.52  & 12.21 & 9.20     & 9.47     & 7.27      & 14.78 & 8.40 & 12.98  \\
\bottomrule
\end{tabular}
}
\vspace{-5mm}
\end{table*}

\noindent\textbf{Merge Layer, Decoder, \& Generation.}
As shown in Fig.~\ref{figure:model}, the representations returned by hybrid bi-attention on each encoder is fused by passing them through a multi-layer perceptron layer. 
Then the decoder concatenates the output and uses attention-based GRU to generate hashtag as a word sequence $\mathbf{y}$.
These parts of model follow~\cite{seo2016bidirectional}.

We follow the previous work~\cite{wang2019microblog} and use negative log-likelihood loss while training. Stochastic Gradient descent algorithm is used to minimize the loss function and learn parameters.
During the inference and generation of hashtags, beam search is applied on the word distribution to select word at each timestamp. We generate a ranking list of hashtags during inference.

\section{Experiments}
\subsection{Datasets}
\noindent\textbf{Tweets.} Two large-scale datasets from English Twitter are collected using Twitter developer API: (1) \emph{Tweets2018}, contains posts published from October 1 to October 13, 2018; (2) \emph{Tweets2020}, contains posts published from May 25 to June 8, 2020.

\noindent\textbf{News.} We crawled news from over 170 newspapers, which were published during September 27 to October 13, 2018 for \emph{Tweets2018} and during May 20 to June 8, 2020 for \emph{Tweets2020}. Nearly 5,000 news articles are collected for each day.

We filter out the posts without any non-inline hashtags\footnote{An inline hashtag is part of a sentence and when segmented into sequence of words, it adds meaning. A non-inline hashtag is usually at the end of the post.}. 
To preserve information in tweets, we replace inline hashtags and mentions in the posts with their respective segmented word sequences using ekphrasis~\cite{baziotis-pelekis-doulkeridis:2017:SemEval2}.
We refer non-inline hashtags as hashtags. Hashtags are segmented into sequences of words and are used as the targets to be generated. 
We detect named entities in the processed posts and news using Stanford CoreNLP toolkit~\cite{manning-EtAl:2014:P14-5}.
 Table~\ref{tab:statistics} summarizes the statistics of the datasets.

\vspace{-2mm}
\subsection{Compared Methods}
We compare with the following methods and we use the same hyperparameters for all generator baselines and our model for fair comparison, and for LSTM-TOP model~\cite{li2016hashtag}, we use the hyperparameters stated in their paper. 
\vspace{-1mm}
\begin{itemize}[leftmargin=*,nosep]
        \item \textbf{Seq2Seq}~\cite{bahdanau2014neural} uses attention based encoder-decoder architecture to generate hashtags from the processed post.
        \item \textbf{Seq2Seq+Copy}~\cite{gu2016incorporating} uses  \textbf{Seq2Seq} architecture with copy mechanism.
        \item \textbf{LSTM-TOP}~\cite{li2016hashtag} is a topic-modeling based approach that uses attention-based LSTM model for learning representations.
\end{itemize}

\textbf{\our-NoRank} is an ablation of our \textbf{\our} that applies BM25 ranking function for retriever instead of our ranking function. 
\textbf{\our-NoRankNoLocal} further replaces our hybrid attention by bi-attention, which is used in~\cite{wang2019microblog}. The implementation details can be found in Appendix C. 

\subsection{Evaluation Metrics and Results}
We adopt some of the evaluation metrics used in \cite{wang2019microblog} including $F_1$ scores at top $K$ predictions ($F_1@K$) where $k = 1, 5, 10$, mean average precision (MAP)~\cite{manning2008introduction} computed on  the top 5 predictions, and ROUGE metric (RG-1)~\cite{lin2004rouge} for the highest-ranked hashtags.  
We also report the prediction accuracy($ACC$) for the highest ranking hashtags. 

Table~\ref{tab:main_result} summarizes the evaluation results of all methods. 
Our proposed framework achieves the best performance among all the compared sequence generation methods.
\our and its variants outperform Seq2Seq and Seq2Seq+Copy with a significant margin.
This observation indicates that news articles are helpful in generating hashtags. The comparison between \our and \our-NoRank demonstrates that the time-aware ranking function which emphasizes emerging entities  retrieves more relevant news, resulting in superior performance.
\our performs better than LSTM-TOP on all metrics on Tweets2018 dataset and on $F_1@1$, accuracy, MAP, RG-1 on Tweets2020 dataset. 
The superior performance is because the topic modeling approaches are extractive in nature and are unable to produce phrasal hashtags. LSTM-TOP slightly outperforms \our on $F1@5$ and $F1@10$ on Tweets2020 dataset. Their model is a classification-based model which selects hashtags from a fix set. For tweets2020 dataset, in the training set,  the top 300 frequent hashtags take up 50\%, leading to the classification-based model to perform well.
\our outperforms the dual encoder with bi-attention model (\our-NoRankNoLocal/~\cite{wang2019microblog}) demonstrating that
the weights of context words reflect their importance and our hybrid attention effectively incorporates it while generating hashtags.

We also present two case studies to demonstrate the effectiveness of \our in Appendix D.

\section{Related Work}
Hashtag Annotation has been modeled as different kinds of problems in the previous works.
Many methods view this task as a classification problem and predict hashtags from a pre-defined list~\cite{zhang2017hashtag, huang2016hashtag, gong2016hashtag}, however, this formulation makes it impossible to generate emerging hashtags.
Another line of approach employs topic models to generate hashtags~\cite{wu2016tag2word,li2016hashtag}. 
However, these models are usually unable to produce phrasal hashtags, which can be achieved by our framework by generating hashtag word sequences.
Since a hashtag summarizes the key ideas of the microblog, some approaches model this problem as keyphrase/sequence generation, for example, using deep RNN model~\cite{zhang2016keyphrase}
and leveraging conversations as a supplementary data source~\cite{zhang2018encoding, wang2019microblog}.
However, it is unrealistic to assume the existence of conversations before the annotation. 
~\cite{sedhai2014hashtag} studies the hashtag recommendation problem for hyperlinked tweets(i.e., tweets containing links to web pages) by mining the context from linked webpages.
In contrast to this, our framework is applicable to any tweet and we leverage widely available, up-to-date news articles to generate hashtags.
Note that, although hashtag generation and text summarization tasks look similar, summarization methods can't be applied to the posts as they are too short and lack enough context to summarize.
Moreover, summarization methods~\cite{nallapati2016abstractive, liu2018generative} usually aim to generate a summary of few sentences, whereas, in the case of hashtags, they have to be short (1-2 words) and carry maximum information to represent the post.

\section{Conclusions and Future Work}
In this paper, we propose \our, a Retriever-Generator framework that leverages news articles published before a microblog post created, to enrich the contextual information and jointly models the context with post to generate hashtags.
Experimental results and case studies demonstrate that our model outperforms previous methods, thereby signifying the advantages of leveraging news articles for generating hashtag.
In the future, we are interested in user-aware hashtag generation by taking the properties of users into consideration.

\bibliography{paper}
\bibliographystyle{acl_natbib}
\newpage\clearpage
\appendix \label{app}
\begin{Large}\textbf{Appendix}\end{Large}
\section{Hashtag Frequency Distribution} \label{app: hashtag distribution }
\begin{figure}[h]
    \centering
    \includegraphics[width=\linewidth]{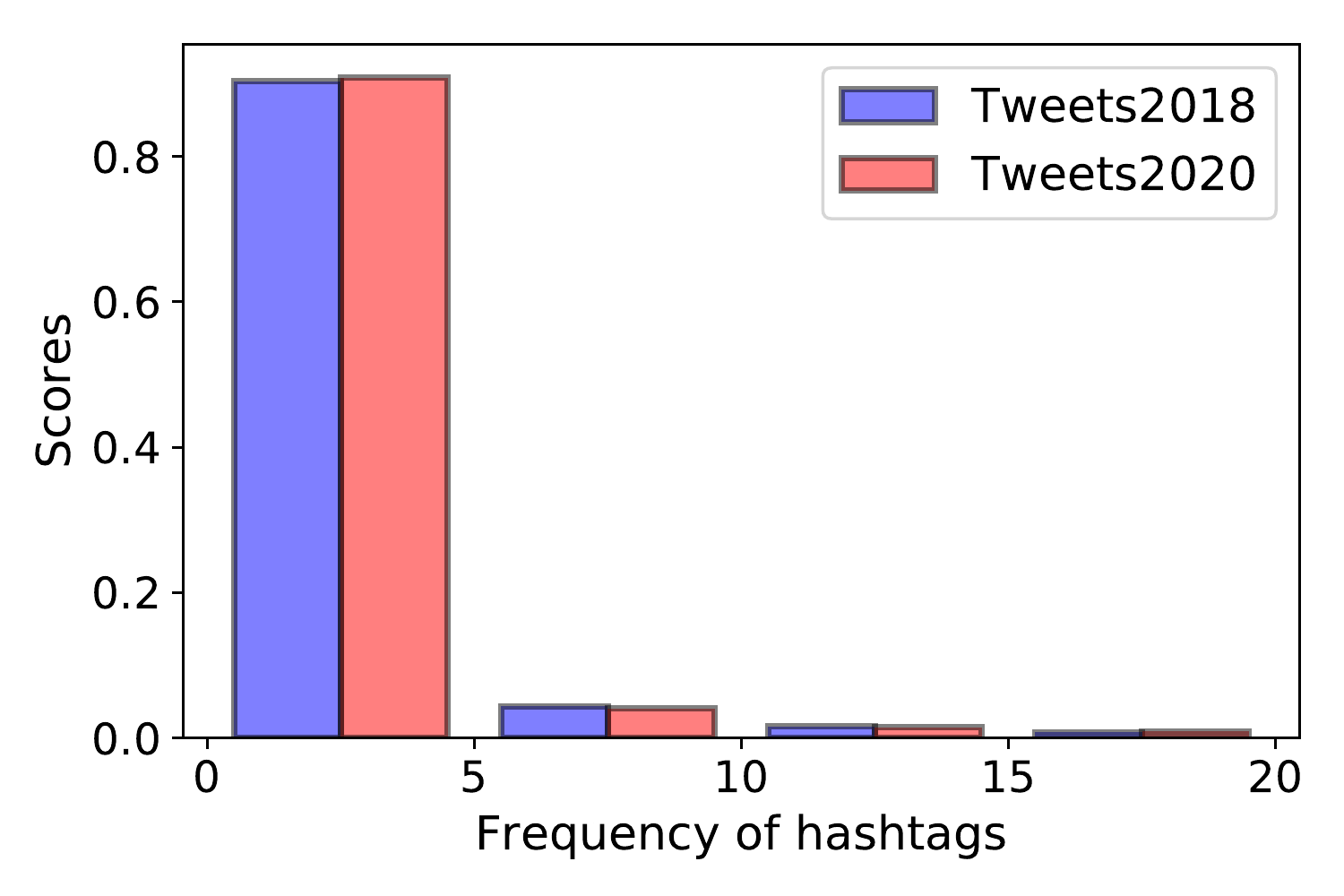}
    \vspace{-3mm}
    \caption{Frequency distribution of hashtags.}
    \label{fig:hashtag-dis}
    \vspace{-3mm}
\end{figure}

The frequency distribution of hashtags in both datasets is shown in Figure~\ref{fig:hashtag-dis}.
This figure shows that the distribution is skewed with more than $80\%$ of hashtags appearing at most $5$ times.

\section{Entity Matching between Microblog and News} \label{app:entity match}


Same entity might be expressed in different ways in microblog posts and news articles.
Therefore, the entities in posts and news have to be carefully matched to better estimate the frequency and IDF of entities in our ranking function.
To address this problem, we develop a soft entity matching mechanism. 


We observe that entities in microblogs are usually written in a concise way, e.g., \textit{Donald Trump} as \textit{Trump}; whereas in news, the formal and complete version of the entity usually occurs at least once in the whole content, and the shorter versions may occur multiple times after it. 
Based on these observations, we propose two intuitive and powerful matching criteria named \emph{Strict Match} and \emph{Conditional Match}.


For a microblog post $p$ and a news article $d \in D_i$, let the entity sequences extracted from them be $E^p$ and $E^d$ respectively.
The strict match aims at finding the formal and longer version of entities in $E^d$ that match to an entity in $E^p$. 

\begin{definition}[Strict Match]~\label{def:strict}
An entity pair $(e^p, e^d)$ is called a strict match if
\begin{equation}\label{equ:match}
\small
  \frac{|\{\mathbf{t}^p | \mathbf{t}^p\in e^p \text{and }\exists \mathbf{t}^d\in e^d \text{ s.t. } align(\mathbf{t}^p, \mathbf{t}^d) \geq t\}|}{|e^p|} \geq q,
\end{equation}
where $\mathbf{t}^p, \mathbf{t}^d$ are tokens of entities $e^p, e^d$, $align(\cdot,\cdot)$ is the character-level alignment score function between two tokens, $|e^p|$ is the number of tokens in $e^p$ and $t,q\in(0, 1]$ are two pre-specified thresholds. 
\end{definition}

According to Definition~\ref{def:strict}, $(e^p, e^d)$ is a strict match when more than $q|e^p|$ tokens in $e^p$ are matched with tokens in $e^d$ at character level, based on alignment score calculated using Smith–Waterman algorithm \cite{smith1981identification}. 

We expand the strict match results set by adding the matching results of shortened versions of strictly matched entities from the same news using our matching rule called Conditional Match, defined as follows:
\begin{definition}[Conditional Match]
For an entity $e^d\in E^d$, entity pair $(e^p, e^d)$ is called a conditional match given $e'^d$ if $\exists$ $e'^d\in E^d$ such that both $(e^p, e'^d)$ and $(e^d, e'^d)$ are strict matches. 
\end{definition}

The following example illustrates the strict match and conditional match.

\begin{example}
Suppose there is a microblog post containing an entity $e_1=\text{``Donald Trump''}$, and a news article containing entities $e_2=\text{``Donald Trump''}$ and $e_3=\text{``Trump''}$. In the first step, $(e_1, e_2)$ is recognized as a strict match since all tokens in $e_1$ are found in $e_2$. $(e_1, e_3)$ is then detected as a conditional match given $e_2$ because (``Trump'', ``Donald Trump'') is also a strict match and $e_2, e_3$ are from the same news article. 
\end{example}

Based on the strict match and conditional match, 
we revise the TF and IDF formulations as follows: 
\begin{equation}\label{equ:tfidf}
    \begin{aligned}
    \small
        f_{e}^{d} &= |M^{soft}(e, d)|,\\
        IDF(e, D_i) &= \log \frac{|D_i|}{|\{d|d\in D_i\text{ and }M^{strict}(e,d)\neq\phi\}|}.
    \end{aligned}
\end{equation}
where
    \begin{equation*}\small
        M^{strict}(e, d) = \{e^d|e^d\in E^d\text{ and }(e,e^d)\text{ is a strict match}\} \\
    \end{equation*}
    \begin{gather*}\small
        M^{soft}(e, d) = \{e^d|e^d\in E^d, \exists e'^d\in M^{strict}(e, d) \text{ and } \\(e^d, e'^d) \text{ is a strict match}\}
    \end{gather*}
\begin{table*}[t]
\footnotesize
    \centering \small
    \begin{tabular}{m{4cm}m{4cm}m{2.5cm}m{2.7cm}m{2.5cm}} 
        \toprule
        \textbf{raw tweet} & \textbf{sentences in relevant news} & \textbf{context words with  weights rank}  & \textbf{our model results}  & \textbf{seq2seq results} \\\midrule
        A huge \#BlackLivesMatter banner has been hung on the fence separating protesters from \textcolor{blue}{Lafayette Park} in front of the \textcolor{blue}{White House} at DC's \#JusticeforGeorgeFloyd \#DCProtests \#WashingtonDCProtests' &  Wednesday got started with law enforcement officers cutting off the main \textcolor{red}{protest} area, near \textcolor{blue}{Lafayette Park} and the \textcolor{blue}{White House}. A large group of protesters laid down on a \textcolor{red}{Washington D.C.} street to repeat the words \textcolor{red}{George Floyd} said as ... & dc: 6th; \newline floyd: 7th; \newline washington:10th; \newline george:13th; \newline protests: 64th & dc protests; \newline dc protest; \newline black lives matter dc; \newline black lives matter; \newline protests 2020 & trump resign now; \newline trumpout2020;\newline trumpresign;\newline defundthepolice;\newline whitehouseprotests  \\\hline
        \textcolor{blue}{Don Lemon} @donlemon Is really shining bright right now! \#BlackLivesMatter \#JusticeForGeorgeFloyd \#GeorgeFloydWasMurdered \#Trump \#TrumpIsAnIdiot \#TrumpIsALoser \#TrumpIsRacist' & 
         ``... Not one of them tried to do anything to help him,'' Tera Brown, \textcolor{red}{Floyd}'s cousin, told CNN's \textcolor{blue}{Don Lemon}.
         Protesters marched in Minneapolis Tuesday night in response to the death of \textcolor{red}{George Floyd}. ... called for the four officers to face \textcolor{red}{murder} charges. 
         & floyd: 1st \newline george: 10th \newline murder: 125th \newline racist: 131st
         &george floyd;\newline police brutality;\newline black  lives matter;\newline george floyd protests;\newline trump & colin powell; \newline message; \newline r4 today;\newline tucker;\newline arizona\\
        
        \bottomrule
    \end{tabular}
    \vspace{-2mm}
    \caption{Case study 2. Two real examples of tweets, news and generated hashtags.}\label{tab:case_study}
\vspace{-2mm}
\end{table*}
\section{Implementation Detail of Models} \label{append: model}
Our generator is implemented on the OpenNMT framework~\cite{klein2017opennmt}.
We choose the embedding size as 300; We use two-layered Bi-GRU cells for encoder and one layer of GRU cells for decoder and the hidden size of GRU is set to 400. 
Adam optimizer~\cite{kingma2014adam} is used to learn parameters with learning rate initialized to $0.001$.
The learning rate decreases with a decay rate of 0.5 and we adopt the early stopping strategy.
The batch size is $64$ and dropout rate is $0.1$. While generating hashtags, we set a maximum sequence length as 10 and the beam size as 20. 

\section{Case Study} \label{sec:case study}
We present two case studies demonstrating the effectiveness of \our.

\begin{figure}[t]
    \centering
    \includegraphics[width=0.45\textwidth]{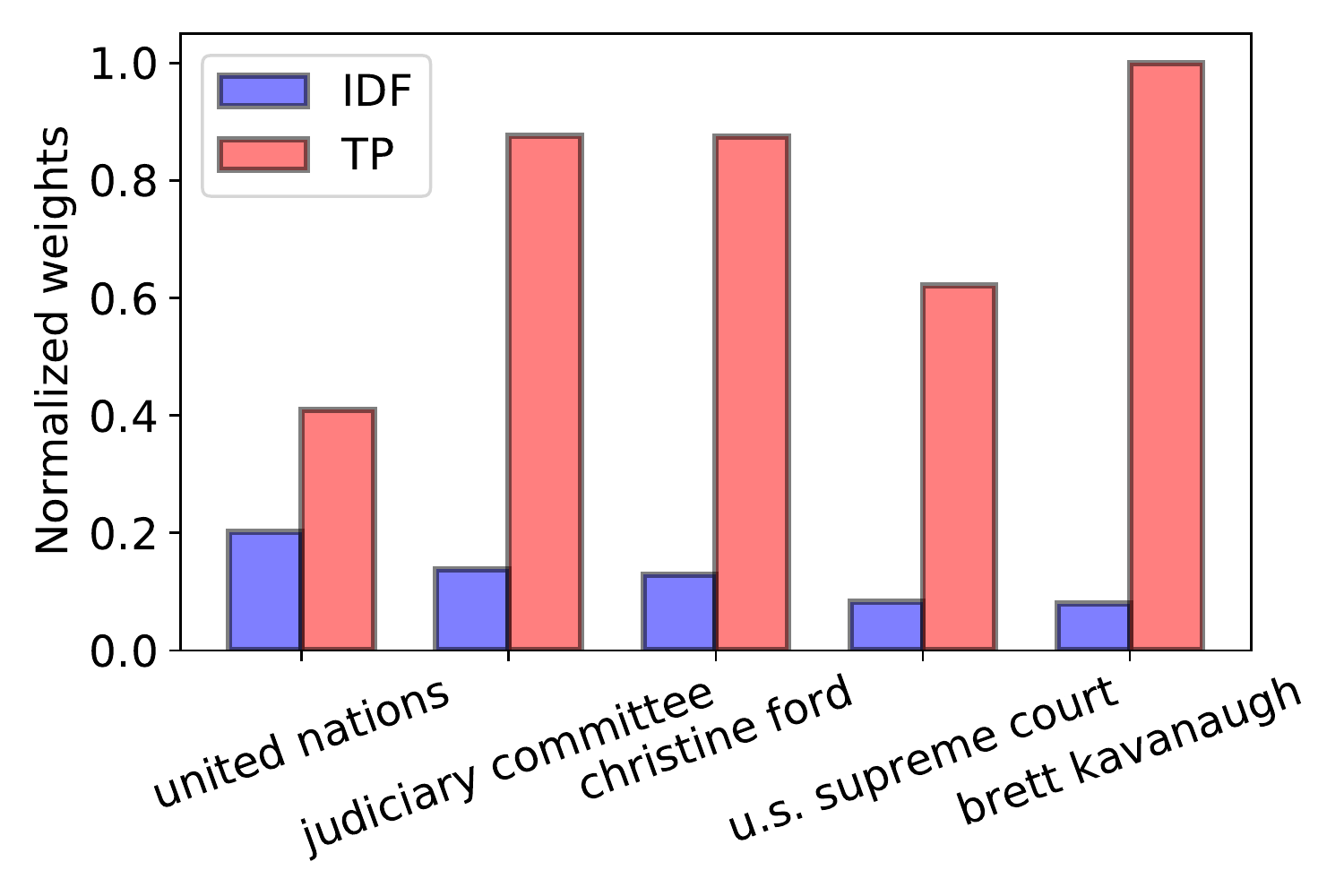}
    \vspace{-3mm}
    \caption{Case study 1. Comparison between IDF score and Temporal Popularity (TP) used in our time-aware entity-focused ranking function. Note that all the values are normalized to range $[0,1]$ for a fair comparison.}
    \label{fig:weights_visualize}
    \vspace{-5mm}
\end{figure}

\noindent\textbf{Effectiveness of time-aware ranking function.}
In BM25, the weight of a query keyword is measured by IDF whereas in our ranking score formulation, it is measured by Temporal Popularity (TP). 
We present the normalized IDF and TP values for a few words from Tweets2018 dataset in Figure~\ref{fig:weights_visualize}.
After going through the news, we learnt about two major events that happened during the time window of the dataset: (1) the 73rd session of the United Nations General Assembly was opened and plenary sessions were going on; (2) the Senate voted 50–48 to confirm Brett Kavanaugh's nomination to the Supreme Court. 
 From Figure~\ref{fig:weights_visualize}, we observe that TP formulation  assigns the emerging and trending entities higher weights compared to BM25 formulation.

\noindent\textbf{Effectiveness of \textbf{\our} on short posts.}
Table~\ref{tab:case_study} presents two examples of real-world tweets, a few sentences from their respective retrieved news, context tokens generated by the retriever, and top hashtags generated by \our and Seq2Seq.
Named entities detected from tweets are colored blue and words in hashtag sequences that appear in news are colored red. 
In the first example, based on the two entities detected in the tweet, news  about the Washington DC protests and George Floyd protests are retrieved which reflect the topic of this tweet.
We observe that many hashtag words can be  found in the context and they have relatively high weights, demonstrating the effectiveness of our time-aware entity-focused ranking function.
The hashtags generated are quite relevant to the tweet and this shows the effectiveness of our generator in leveraging the news articles during generation.
In the second example, since the tweet is really short, without much contextual information, the Seq2Seq generates irrelevant hashtags. 
However, our framework selects news  related to George Floyd's death using the entity ``Don Lemon'' and generates hashtags that are informative and relevant to the tweet. 
\end{document}